\documentclass[a4paper, 10pt, conference]{ieeeconf}      

\IEEEoverridecommandlockouts                              

\usepackage{hyperref}
\usepackage{graphicx} 
\usepackage{color}
\usepackage{fullpage} 
\usepackage{float} 
\usepackage{caption}
\usepackage{subcaption}
\usepackage{amsmath}
\usepackage{amssymb}
\usepackage{booktabs}
\usepackage{adjustbox}
\usepackage{algorithm2e}
\usepackage[noend]{algpseudocode}
\usepackage{cite}

\graphicspath{{figures/}}
\title{\LARGE \bf
Mobility Map Computations for Autonomous Navigation using an RGBD Sensor}

\author{Nicol\`o Genesio$^\dag$, Tariq Abuhashim$^\dag$, Fabio Solari$^\ddag$, Manuela Chessa$^\ddag$ and Lorenzo Natale$^\dag$ 
\thanks{This research has received funding from the European Union's Seventh Framework Programme for research, technological development and demonstration under grant agreement No. 611909 (KoroiBot).}
\thanks{$^\dag$ are with the iCub Facility, Istituto Italiano di Tecnologia, via Morego, 30, 16163 Genova, Italy email:
        {\tt\small \{nicolo.genesio, tariq.abuhashim, lorenzo.natale\}@iit.it}}%
\thanks{$^\ddag$ are with the DIBRIS, Universit\`a degli studi di Genova, Via All'Opera Pia, 13, 16145 Genova, Italy email:
        {\tt\small \{fabio.solari,manuela.chessa\}@unige.it}}%
}

\begin{document}

\maketitle
\thispagestyle{empty}
\pagestyle{empty}

\begin{abstract}

In recent years, the numbers of life-size humanoids as well as their mobile capabilities have steadily grown. Stable walking motion and control for humanoid robots are active fields of research. In this scenario an open question is how to model and analyse the scene so that a motion planning algorithm can generate an appropriate walking pattern.

This paper presents the current work towards scene modelling and understanding, using an RGBD sensor. The main objective is to provide the humanoid robot iCub with capabilities to navigate safely and interact with various parts of the environment. In this sense we address the problem of traversability analysis of the scene, focusing on classification of point clouds as a function of mobility, and hence walking safety.


\end{abstract}

\section{Introduction}

The potential market of service and entertainment humanoid robots has attracted great research interests. One of the most fundamental and challenging steps is to allow robots to interact and walk autonomously within a real world scenario.
Ideally, they should be able to accept high level human commands and, then, autonomously walk in a real-life environment consisting of floors and stairs without colliding with obstacles. Thus, it is necessary for humanoids to be able to identify candidate traversable places within acceptable stability limits. This includes, for instance, flat surfaces or surfaces with limited slope and roughness values.

\begin{figure}[t] 
	\centering
        \includegraphics[width=0.5\textwidth]{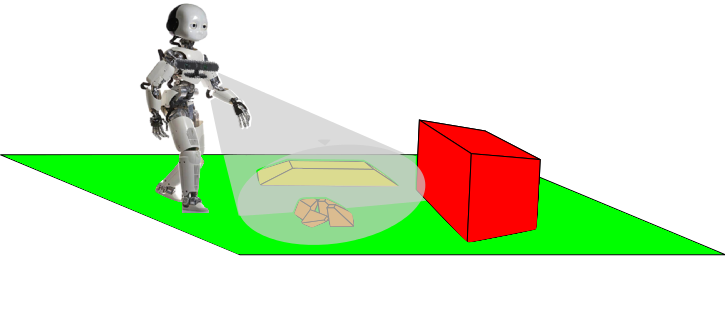}
        \caption{Example of what we want to obtain, a map where each region has a score between 0(red) and 1(green) that indicates how much traversable a surface is.}
\vspace{-0.6 cm}
\label{fig:mobility}
\end{figure}

This paper presents a pipeline to compute a mobility map in real-time using an RGBD sensor mounted on the iCub, as shown in Figure \ref{fig:mobility}. 
As in most navigation scenarios the terrain information is vital to classify what is traversable and what is not.
In particular, legged-type robots such as humanoids require precise information on the surrounding terrain, not only for determining locomotion strategies, but also for coordinating complex body motions, such as generating obstacle-avoiding free-leg trajectories during walking. 


For humanoids, and from a simplified kinematic point of view, a surface is considered as a part of an obstacle if its geometric properties do not allow for a safe foot step planning. Useful surface properties may include its slope, roughness, rugosity and size. Such properties can be measured or estimated from 3D data, which can be provided, for instance, by using a structured light sensor. Such sensors are considered cheaper and lighter than their counterpart laser scanners. They also provide richer visual information, which can be useful for tasks including object recognition, classification and segmentation.

\section{Related Work}

In recent years, the interest in scene modelling and understanding algorithms has widely spread. Specifically, segmentation and classification of outdoor workspaces using 3D depth data \cite{Dima3}, in order to allow mobile robots to move in real world environments while navigating between obstacles safely.\cite{Ferri10}.
%

The literature is rich also of research examples that deals with the problem of indoor scene understanding in order to give to the humanoid robot the capability of moving in autonomy in a domestic environment \cite{Li(b)}.
The terrain plays a key role in works like this, and an overview on terrain traversabilty methods can be found in  \cite{Papadakis}. Most of the methods use 3D geometric features (\textit{e.g.} normals)
to interpret the scene that in general are very computationally demanding. For this reason these approaches are not appropriate to manage dynamic obstacles, for which the map has to be updated frequently and then the computation has to be repeated several times. On the other hand approaches that \textit{learn} the scene through proprioceptive informations are more suitable for online applications. This approach is demonstrated
in \cite{Rover} for a Mars-Rover platform where the authors represent
the environment in proprioception space in terms of expected
slip.
Moreover proprioceptive informations can be used in addition of 3D data to increase the reliability of the slope inference and then the accuracy of the classification \cite{KarumanchiABS7}.

More in general the reliability in obstacle detection and scene classification can be augmented relying on multiple sensing modalities such as color and 3D data \cite{Ras4}. 
Visual and structural modalities are clearly complementary: vision alone may be inadequate or unreliable in the presence of strong shadows,
while depth measurement of sensors like structured light sensors can be misled by the sunlight interference with the IR sensor.
In literature can be found works that try to solve the same problem, but using different tools to obtain this kind of information as laser range-finder \cite{Ras4}, or  stereo vision \cite{Takaoka(h)}, but vision techniques are in general challenged by darkness or other extreme lighting conditions and the accuracy that they can provide is too low to manage a such complex issue as scene modelling and interpretation.

\section{Our Implementation}

\begin{figure}[t] 
	\centering
        \includegraphics[width=0.4\textwidth]{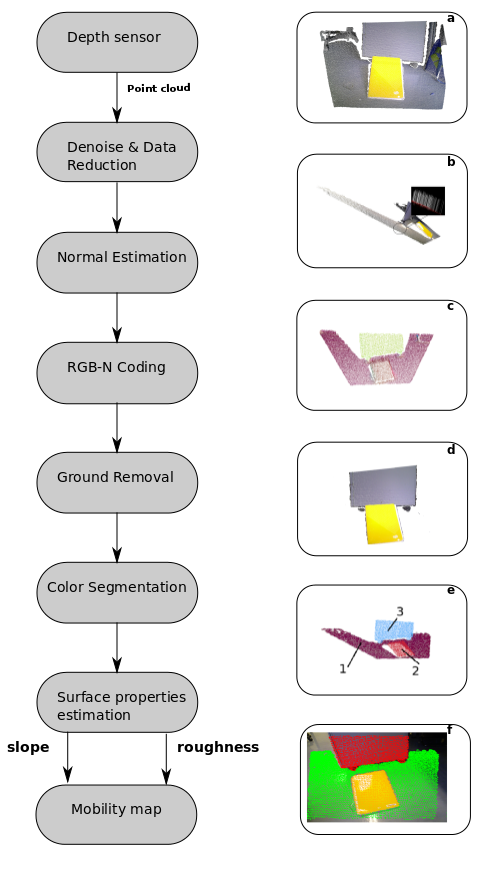}
        \vspace{-0.4cm}
        \caption{Block diagrams displaying the main points of our algorithm.}
\vspace{-0.3cm}
\label{fig:pipeline}
\end{figure}

This section describes our mobility map implementation, where Figure \ref{fig:pipeline} highlights different data processing and estimation steps.

\textbf{Data reduction}: 
In this paper, we used the Asus Xtion PRO LIVE. This sensor produces depth maps at 30 fps with a resolution of 640$\times$480. The depth map includes over 300 thousands of points. In the next steps, it is required to evaluate a surface properties at any point of interest using its $k$-nearest neighbours. The computation cost is, hence, $\mathcal{O}(nk)$, where $n$ is the number of points in the depth map. Based on the application, the way the sensor is mounted on the robot and the sensor field of view, we are able to reduce the computational cost of the depth map by downsampling it into voxels of 1cm$\times$1cm$\times$1cm. Further data reduction can be achieved by limiting the depth range of the sensor to include only points with acceptable depth uncertainty. In this paper, we removed points at distances longer than 1.5-2 meters.

\textbf{Denoising}: 3D information contained in the point cloud is often contaminated by noise. This is due to many causes including the interference of daylight with the IR sensor and intrinsic errors made by the sensor performing the triangulation. Before attempting to estimate the characteristics of a point with respect to its surrounding, it is important to analyse if the surrounding neighbourhood is a good representation of the underlying sampled surface. Thus, following the method proposed in \cite{Rusu08RAS2}, for each point $p_q$ in a cloud $P$, the mean distance $\bar d$ to its $k$-nearest neighbours is first computed. Then a distribution over the mean distance space for the entire point cloud is assembled and its mean $\mu_k$ and standard deviation $\sigma_k$ are estimated. Our motivation is to keep points with mean distance $\bar d$ to the $k$-nearest neighbours that is statistically similar to those by rest of the points. Thus, the new down-sampled point cloud $P^*$ can be extracted using,
\begin{equation}
\label{eq:statistical}
P^*=\{ p_q^* \in P | ( \mu_k - \alpha\sigma_k ) \le \bar d \le ( \mu_k + \alpha\sigma_k )\}
\end{equation}
where $\alpha$ is a desired density restrictiveness factor.

\textbf{Normal Computation}: 
For a good representation and segmentation we have to use features with high discriminating power such as normals and curvature.
They are two of the most widely used geometric features because they provide information on the orientation of surfaces, indispensable for scene understanding.
They are treated as local features, because they characterize the information provided by the $k$-nearest neighbours of each point. Their estimated values are sensitive to sensor noise and the selection of the $k$ neighbours.

After we have determined the neighbourhood $P^k$ of a query point $p_q$, we can use it to compute a local feature that represents the geometry around the query point. One surface point feature can be computed as the normal vector of the tangent plane, which can be estimated solving a least-square plane fitting problem over $P^k$ \cite{shakarji1998least}.
As explained in \cite{RusuDoctoralDissertation} this plane can be represented by a point \textit{x} and a normal vector $\vec{n}$, and, given the distance from a point $p_i \in P^k$ to the plane as $d_i = ( p_i - x ) \cdot \vec{n}$, the values of \textit{x} and $\vec{n}$ are computed in a
least-square sense such that $d_i = 0$.

Assume
\begin{equation}
\label{eq:centroid}
x = \bar{p} = \frac{1}{k} \sum_{i=1}^k p_i
\end{equation}
as centroid of $P^k$, we can then solve the fitting problem obtaining $\vec{n}$ as the eigenvector $\vec{v_0}$ of the smallest eigenvalue $\lambda_0$ of the covariance matrix $C \in R^{3x3}$ of $P^k$, expressed as
\begin{equation}
\label{eq:cov}
C=\frac{1}{k} \sum_{i=1}^k (p_i - \bar{p}) \cdot (p_i - \bar{p})^T.
\end{equation}
To resolve for the sign of $\vec{n}$, however, we need the viewing point of the sensor $v_p$. To define the sign of the normals, each $\vec{n_i}$ has to satisfy
\begin{equation}
\label{eq:sign}
\vec{n_i} \cdot (v_p-p_i) > 0. 
\end{equation}
In this way all the normals $\vec{n_i}$ point towards the viewpoint of the sensor.

\textbf{RGB-N coding}: Before segmentation, we colour-code the point cloud using the computed normals. In the previous step, we have computed, for each point $p_i$, a normal $\vec{n_i}=\{n_{ix}, n_{iy}, n_{iz}\}$, function of its neighbourhood. We assign to $p_i$ the following RGB-coding
\begin{equation}
\begin{split}
R &= \frac{255}{max_{nx}-min_{nx}}*(n_{ix}-min_{nx}) \\
G &= \frac{255}{max_{ny}-min_{ny}}*(n_{iy}-min_{ny}) \\
B &= \frac{255}{max_{nz}-min_{nz}}*(n_{iz}-min_{nz}),
\end{split}
\end{equation}

where $max_{n*}$ and $min_{n*}$ are respectively the maximum and minimum value of the component $n*$ among all the normals contained in the cloud. By doing so, point-cloud's surfaces with similar normal, and thus orientation, will have consistent and similar colors (Fig. \ref{fig:pipeline}.c).
This is an important step which allows us to use color differences to segment regions with common geometric features.

\textbf{Ground removal}: Ground removal is an important step for robust segmentation results. It provides two main benefits: First, it has the effect of isolating the object from the background, and thus improving segmentation. Second, it allows us to further reduce the computational cost of the entire system, because we exclude from the analysis the segment that has the majority of the points in the cloud.

There are many ways to segment the floor from the rest of the scene, some of them use images \cite{GroundImages2} but, despite their robustness, they are often afflicted by the differences in light conditions of the scene. One way to achieve robustness is by using RANSAC, using the following steps:
\begin{enumerate}
\item Randomly select three non-collinear unique points $\{p_i , p_j , p_k \}$ from the point cloud $P$ ;
\item Compute the model coefficients from the three points $(ax + by + cz + d = 0)$;
\item Compute the distances from all $p \in P$ to the plane model $( a, b, c, d )$ ;
\item Count the number of points $p^* \in P$ whose distance d to the plane model falls between
$0 \le | d | \le | d_t |$ , where $d_t$ represents a user specified threshold.
\end{enumerate}
The last step represents a way of ``scoring'' a specific model that we used to find the best plane in the cloud. Every set of points $p^*$ is stored, and the above steps are repeated for $k$ iterations. The number of iterations is defined as follow. If $\epsilon$ is the probability of picking a sample that produces a bad estimate (\textit{i.e.} outlier), then $1-\epsilon$ is the probability of picking at least one good sample (\textit{i.e.} inlier). This means that the probability of picking $\gamma$ good samples becomes $( 1 - \epsilon)^\gamma$. For $k$ trials, the probability of failure becomes
$( 1 - ( 1 - \epsilon )^\gamma )^k$. If $p$ is the desired probability of success (\textit{e.g.} $p$ = 0.99), then:
\begin{equation}
\label{eq:GrndK}
1 - p = ( 1 - ( 1 - \epsilon )^\gamma )^k
\implies 
k =
\frac{\log{( 1 - p)}}
{\log{( 1 - ( 1 -\epsilon )^\gamma )}}.
\end{equation}
After the algorithm is terminated, the set with the largest number of points (inliers) is selected as the support for the best planar model found, the ground in our case.

\textbf{Color segmentation}: 
After removing the ground, our objective is to segment the remaining regions in the cloud. A RANSAC-based method wont suffice for this task, because it assumes that all the objects in the scene can be mathematically modelled. Thus, we utilize \textit{Region Growing} \cite{Adams:1994:SRG:628313.628615} for this task. This method follows
a \textit{flood fill} approach. The method aims at selecting a set of {\it homogeneous} points by optimising for a given inter-regional constraint. A \textit{seed} point is first selected. At each optimisation step, the surrounding of the seed point is iteratively allowed to \textit{grow} by including more points into the computations of the local constraint. This process is iterated until this local constraint is satisfied. Typical constraints may include the Euclidean distance between the selected point and the seed point or local features, including geometric and photometric features. For more efficient computations, we utilise the previously color-coded point-cloud based on the computed surface normals as a feature, and formulate our regional constraints using the Euclidean distance. This choice simply allows us to extract regions with homogeneous surface orientations. 
Algorithm \ref{alg::Reggrow} summarizes steps of Region Growing on 3D point-clouds, where
\begin{algorithm}[tbp]
\While{$\{A\}$ is not empty}{
	Current Region $\{R_{c}\} \leftarrow \oslash$;\\
	Current seeds $\{S_{c}\} \leftarrow \oslash$;\\
	Select randomly $p$ from $\{A\}$;\\
	$\{S_{c}\} \leftarrow \{S_{c}\} \bigcup p $; $\{R_{c}\} \leftarrow \{R_{c}\} 	\bigcup p $;\\
	$\{A\} \leftarrow \{A\} \setminus p $;\\
	\For{i=0 to size($\{S_{c}\}$)}
		{Nearest neighbours of p: $B_{c}\leftarrow\Omega(\{S_{c}\})$;
		\For{j=0 to size($\{B_{c}\}$)}
		{
			Current neighbour point $p_{j}$;\\
			\If{$\{A\}$ contains $p_{j}$ and $coldist_{pj}<c_{th}$ }{
				$\{S_{c}\} \leftarrow \{S_{c}\} \bigcup p_{j} $; $\{R_{c}\} 								\leftarrow \{R_{c}\} 	\bigcup p_{j} $;\\
				$\{A\} \leftarrow \{A\} \setminus p _{j}$;}
		
		}
		}
		$\{R\} \leftarrow \{R\} \bigcup \{R_{c}\}$;
		
}
\caption{Region Growing based on color coding of surface normals.}
\label{alg::Reggrow}
\end{algorithm}
$P$ is a color-coded point-cloud, $c_{th}$ is a color threshold, $coldist_{pj}$ is the computed Euclidean distance between points $p$ and $p_{j}$ in the color space ($YC_{r}C_{b}$), $R$ is the region list, $\Omega(.)$ is the neighbour finding function, and $A$ is the available point list, that it is initialized using all the points of $P$.

\textbf{Surface properties estimation}: After dividing the cloud in clusters with similar geometric properties (Fig. \ref{fig:pipeline}.e), we estimate those properties that will be used to determine the mobility scoring of individual segments in the point-cloud. First, we compute the slope, which is a measure of steepness for planes or more in general for flat regions. First, we estimate the best plane that fits the points within the \textit{i-th} segment using least-squares and RANSAC. The slope $s_i$ is then calculated as the angle between the fitted plane normal $\vec{n_i}$ and the vector $\vec{n_{f}}$ representing the normal vector of the removed ground segment. This can be simply computed as:
\begin{equation}
\theta = arcos(\frac{\vec{n_i} \cdot \vec{n_f}}{||\vec{n_i}|| \cdot ||\vec{n_f}||}) * \frac{180}{\pi}.
\label{eq:slope}
\end{equation}
In addition to the slope, our mobility function considers the surface roughness, which represents as a measure of the asperity of a certain surface. Surface roughness is widely used in mobile robotics navigation planning \cite{shockparam13}, because of its ability to limit the mobility of various robotic platforms. For humanoids, safe and stable foot step planning is highly correlated to terrain roughness.

There are many ways to estimate the roughness mostly because a canonical definition has not already formulated. We followed an approach already used in\cite{Friedman9}. For each segment of the cloud we compute the roughness index $r$ as:
\begin{equation}
\label{eq:rough1}
r=\frac{A}{A'},
\end{equation}
where $A$ is the area of the segment, and $A'$ is the area of the segment projection on the corresponding estimated plane, which was computed using RANSAC and least-squares. To compute $r$ we utilize \textit{Delaunay triangulation}. Given a set of vertices $V = \{v_m\}$, such that $v_m \in \mathbb{R}^3$ and $m = 1, ..., MS$, and $v_m = (x_m , y_m , z_m )$ represents the vertex $m$ described by its $x, y, z$ point coordinates. The triangles of the surface are contained in the set $T = \{t_n \}$, where $n = 1, ..., N$, such that $t_n \subset V$ and $t_n = (v_{1_n} , v_{2_n}, v_{3_n} )$ represents a triangle defined by three vertices in $V$. Thus, $A$, and similarly $A'$, are computed using:
\begin{equation}
\label{eq:rough2}
A=\sum_{j=1}^{N}a_j,
\end{equation}
where $a_j$ is the area of the \textit{j-th} triangle ($t_j$), which is computed as half the magnitude of the cross product of the vectors ($\vec{v_{1}v_{2_j}}$ and $\vec{v_{2}v_{3_j}}$) representing two adjacent sides of the triangle. Thus,
\begin{equation}
\label{eq:rough3}
a_j=\frac{1}{2} ||\overrightarrow{v_{1}v_{2_j}}\times\overrightarrow{v_{2}v_{3_j}}||.
\end{equation}

\begin{figure}[h] 
	\centering
        \includegraphics[width=0.5\textwidth]{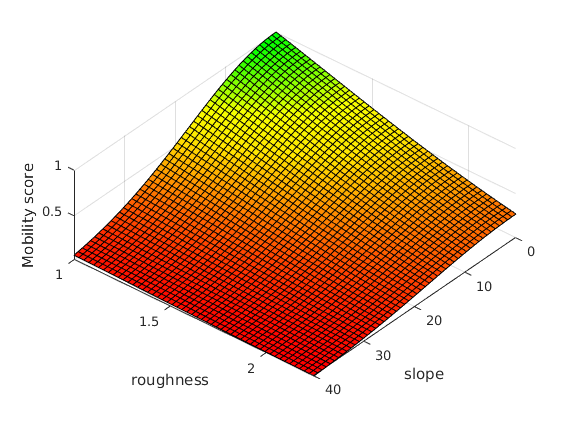}
        \caption{A sample mobility cost function, which was generated with GPR.}
\label{fig:rule}
\end{figure}

\textbf{Mobility map}:
The classification of the scene as a function of robots mobility is performed using a simple and intuitive approach. Every segment is assigned a score $\mathbf{y}$ between 0 and 1, where 0 corresponds to \textit{untraversable} surfaces while 1 corresponds to \textit{traversable} surfaces that does not require any additional foot step planning, which is shown in table \ref{tab:rule}. This mobility rule is very discrete, and hence limits its applications to multiple foot step configurations. In order to accommodate finer mobility decisions, we further improve the resolution of our decisions on the traversability of surfaces by applying Gaussian Process Regression (GPR) with a Squared Exponential (SE) covariance function \cite{Rasmussen:2005:GPM:1162254} (see Fig. \ref{fig:rule}):
\begin{equation}
k_y(x_i,x_j) = \sigma_f^2\exp\left(-\frac{1}{2\lambda^2}(x_i-x_j)^2\right),
\end{equation}
where $(x_i,x_j)$ are input pairs containing surface features, namely the estimated slope and roughness, $x_i=(s_i,r_i)^\top$, $\sigma_f$ is the variance (we assumed, $\sigma_f=1$) and $\lambda$ is the length scale which defined the smoothness of the mobility score (we assumed, $\lambda=0.1$). Since we would like our mobility to be a function of both slope and roughness, we train our GPR model using two the inputs ($s$ and $r$) with one output mobility score $m$. Thus, we can obtain the score as an inference instance using our trained GPR model:
\begin{equation}
\hat{m}(x_t) = \mathbf{k}_*^\top(\mathbf{K}+\sigma_n^2\mathbf{I})^{-1}\mathbf{y},
\end{equation}
where, $\mathbf{K}$ is the covariance matrix, $\mathbf{k}_*$ to denote the vector of covariances between the test point $x_t=(s_t,r_t)^\top$ and the training points $x_i=(s_i,r_i)^\top$, $\sigma_n$ is the expected noise in the measured mobility $\mathbf{y}$, given the normals.
The range of input dimensions was limited by the walking capabilities and mechanical limits of the robot.
\begin{table}[h]
\centering
\begin{adjustbox}{width=0.4\textwidth}
\label{my-label}
\begin{tabular}{@{}l|rrrrr@{}}
\toprule
\textbf{\begin{tabular}[c]{@{}l@{}}s[$^{\circ}$]$\rightarrow$ \\ r$	\downarrow$\end{tabular}} & 0-10 & 10-20 & 20-30 & 30-40 & \textgreater40 \\ \midrule
1-1.2                                                                         & 1    & 0.75  & 0.5   & 0.25  & 0              \\
1.2-1.4                                                                       & 0.75 & 0.5   & 0.25  & 0     & 0              \\
1.4-1.6                                                                       & 0.5  & 0.25  & 0     & 0     & 0              \\
1.6-1.8                                                                       & 0.25 & 0     & 0     & 0     & 0              \\
1.8-2.0                                                                       & 0    & 0     & 0     & 0     & 0              \\
2.0-2.2                                                                       & 0    & 0     & 0     & 0     & 0              \\
\textgreater2.2                                                               & 0    & 0     & 0     & 0     & 0             
\end{tabular}
\end{adjustbox}
\caption{Mobility rule estimated empirically.}
    \label{tab:rule}
\end{table}

\section{Experiments}

In this Section we present supporting experimental results to demonstrate the performance of the proposed mobility map computation pipeline. The collected dataset comprised 3D depth maps and associated RGB images. Due to the limited mobility of the iCub robot and the sensitivity of the depth sensor to daylight, we tested our implementation only on data sets collected for indoor environments. For example, laboratory corridors with various objects being located along the way. The method of this paper was implemented using C++, and is available on GitHub\footnote{\url{https://github.com/Nicogene/MobilityMapBuilder}.}.

\subsection{RGBD Sensor Model}
The validation of our system has been made through the projection of the 3D mobility maps on the images acquired by the RGB camera of Asus Xtion Pro Live(Fig. \ref{fig:cleanphotos}, \ref{fig:obstaclephotos}). The mapping has been done using the following pinhole camera model,
\begin{equation}
\label{eq:Sensor Model}
\begin{split}
x &= c_{x} + f_{x}*\frac{X}{Z}-o_x\\
y &= c_{y} + f_{y}*\frac{Y}{Z}-o_y,
\end{split}
\end{equation}
where $\{x,y\}$ are the image coordinate, $\{X,Y,Z\}$ are the 3D point coordinates, $\{c_{x},c_{y}\}$ define the optical center, $\{f_{x},f_{y}\}$ are the focal lengths and $\{o_{x},o_{y}\}$ are the factory offsets between the IR sensor frame and the RGB camera frame.
 
\subsection{Free corridor}

\begin{figure}[h!]
	\begin{subfigure}{0.23\textwidth}
		\centering
		\includegraphics[width=0.95\textwidth]{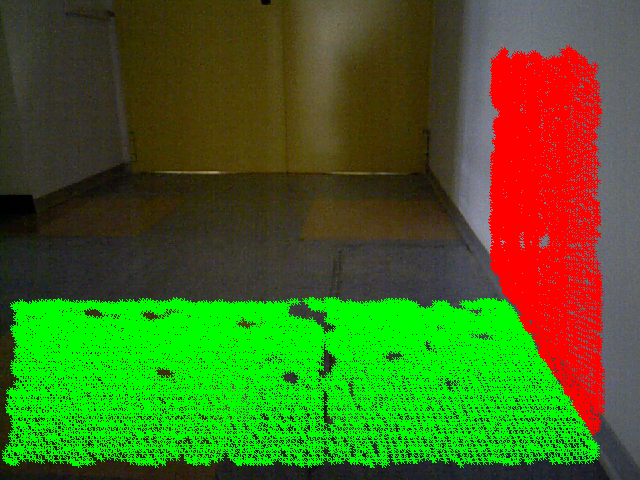}
		\caption{}
		\label{}
	\end{subfigure}
	\begin{subfigure}{0.23\textwidth}
		\centering
		\includegraphics[width=0.95\textwidth]{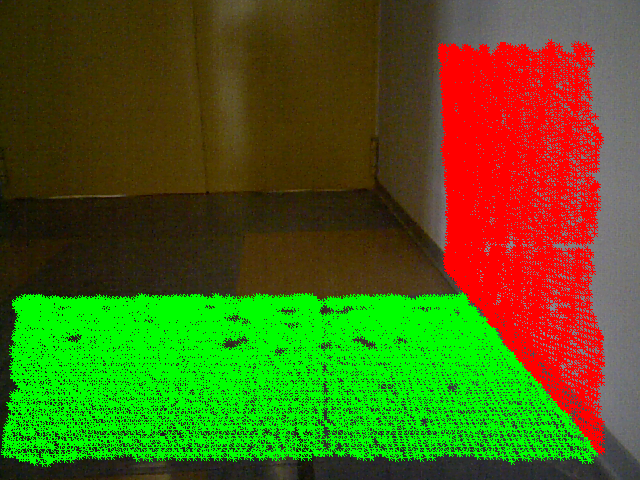}
		\caption{}
		\label{}
	\end{subfigure}\\
	\begin{subfigure}{0.23\textwidth}
		\centering
		\includegraphics[width=0.95\textwidth]{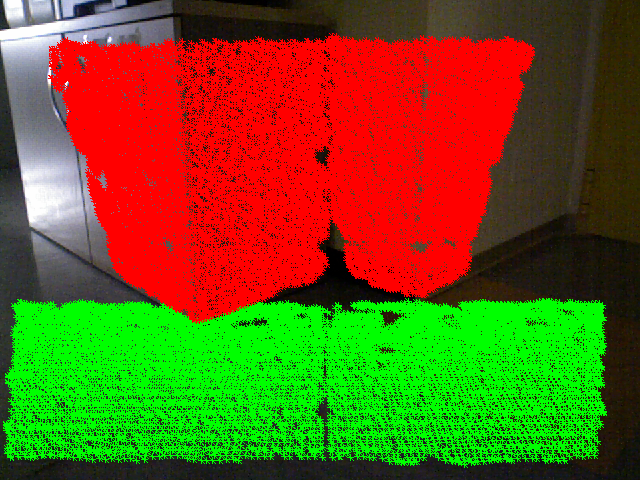}
		\caption{}
		\label{}
	\end{subfigure}
	\begin{subfigure}{0.23\textwidth}
		\centering
		\includegraphics[width=0.95\textwidth]{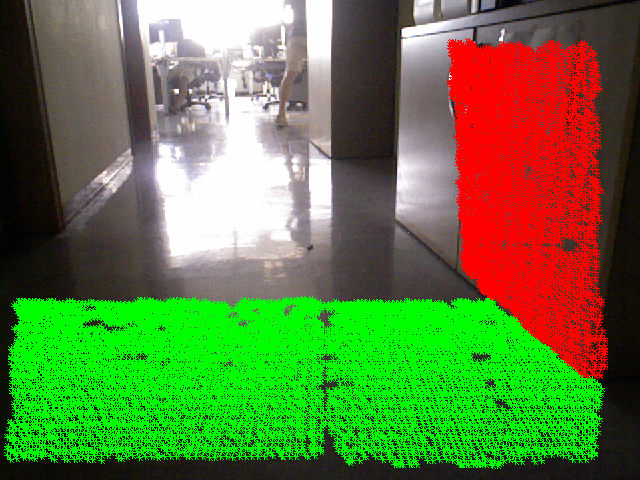}
		\caption{}
		\label{}
	\end{subfigure}\\
\begin{subfigure}{0.23\textwidth}
		\centering
		\includegraphics[width=0.95\textwidth]{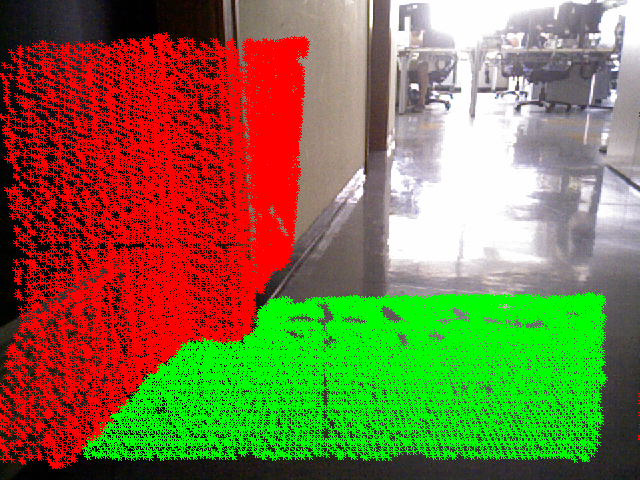}
		\caption{}
		\label{}
	\end{subfigure}
	\begin{subfigure}{0.23\textwidth}
		\centering
		\includegraphics[width=0.95\textwidth]{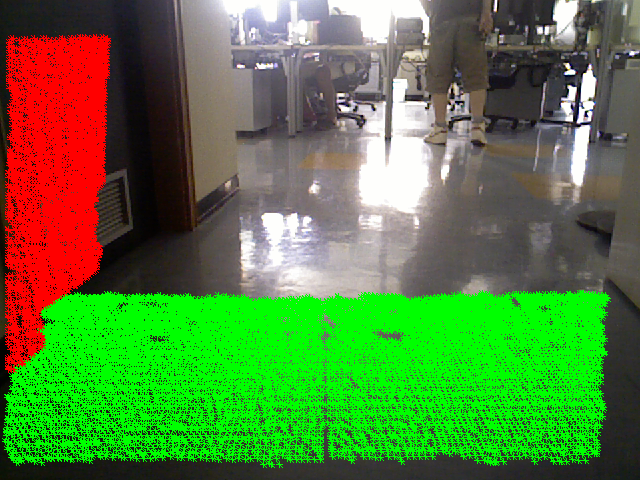}
		\caption{}
		\label{}
	\end{subfigure}
\caption{Images displaying the path crossed, in sequence from (a) to (f). We projected on them the clusters coloured in function of the traversability; we used a color map where green means \textit{mobility score=1} and red means \textit{mobility score=0}.}
\label{fig:cleanphotos}
\end{figure}

\begin{figure}[h!]
	\begin{subfigure}{0.23\textwidth}
	\centering
		\includegraphics[width=0.8\textwidth]{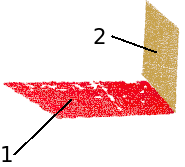}
		\caption{}
		\label{}
	\end{subfigure}
	\begin{subfigure}{0.23\textwidth}
	\centering
		\includegraphics[width=0.8\textwidth]{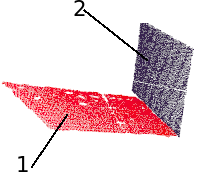}
		\caption{}
		\label{}
	\end{subfigure}\\
	\begin{subfigure}{0.23\textwidth}
	\centering
		\includegraphics[width=0.8\textwidth]{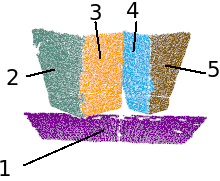}
		\caption{}
		\label{}
	\end{subfigure}
	\begin{subfigure}{0.23\textwidth}
	\centering
		\includegraphics[width=0.8\textwidth]{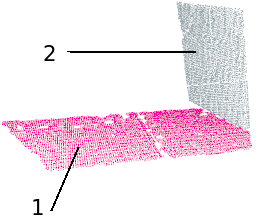}
		\caption{}
		\label{}
	\end{subfigure}\\
	\begin{subfigure}{0.23\textwidth}
	\centering
		\includegraphics[width=0.8\textwidth]{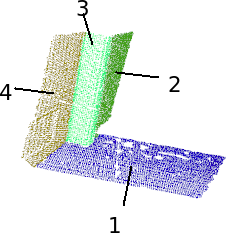}
		\caption{}
		\label{}
	\end{subfigure}
	\begin{subfigure}{0.23\textwidth}
		\centering
		\includegraphics[width=0.8\textwidth]{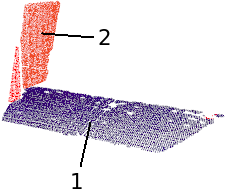}
		\caption{}
		\label{}
	\end{subfigure}
        \caption{Figures presenting point clouds acquired and segmented with our normals/color based method.} 
\label{fig:cleanclus}
\end{figure} 



Fig. \ref{fig:cleanphotos} shows mobility map estimation and point-cloud segmentation results in the image frame using our pipeline for the case of obstacle free path. Our motivation is to initially test our implementation of ground removal, since the performance of our method relies on it. Also, showing the results in the image frame provides us with a way to validate the accuracy of our implementation. On the other hand, Fig. \ref{fig:cleanclus} shows the verification of our segmentation results in 3D space.


\subsection{Corridor with obstacles}

\begin{figure}[h!]
	\begin{subfigure}{0.23\textwidth}
		\centering
		\includegraphics[width=0.95\textwidth]{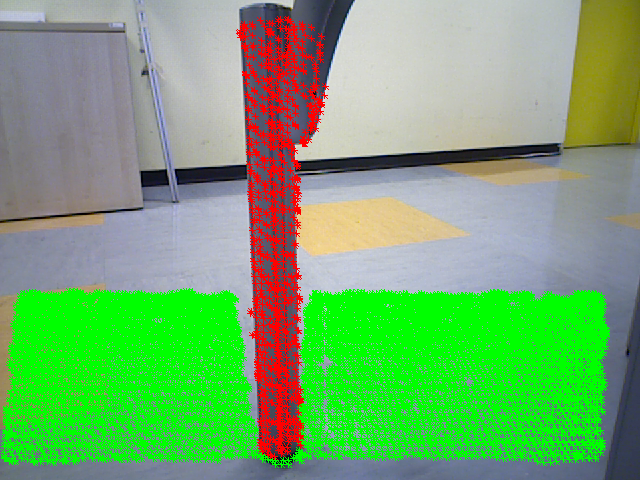}
		\caption{}
		\label{}
	\end{subfigure}
	\begin{subfigure}{0.23\textwidth}
		\centering
		\includegraphics[width=0.95\textwidth]{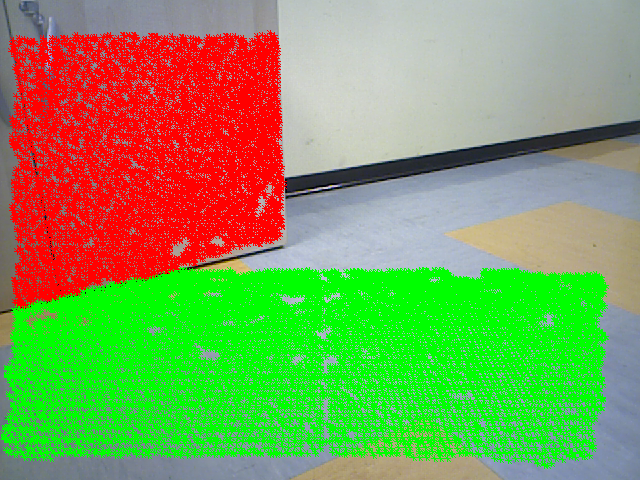}
		\caption{}
		\label{}
	\end{subfigure}\\
	\begin{subfigure}{0.23\textwidth}
		\centering
		\includegraphics[width=0.95\textwidth]{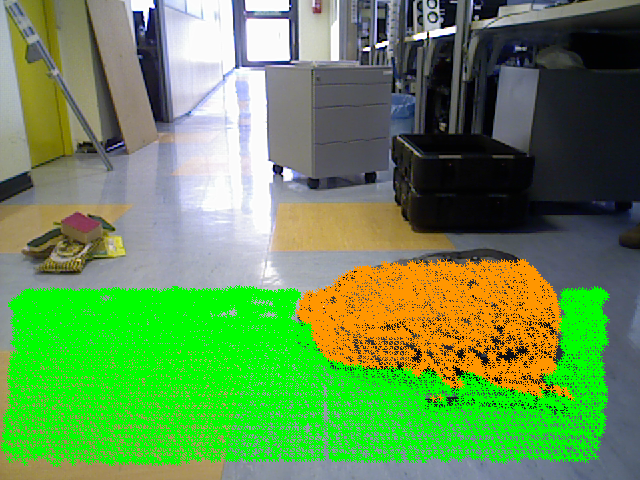}
		\caption{}
		\label{}
	\end{subfigure}
	\begin{subfigure}{0.23\textwidth}
		\centering
		\includegraphics[width=0.95\textwidth]{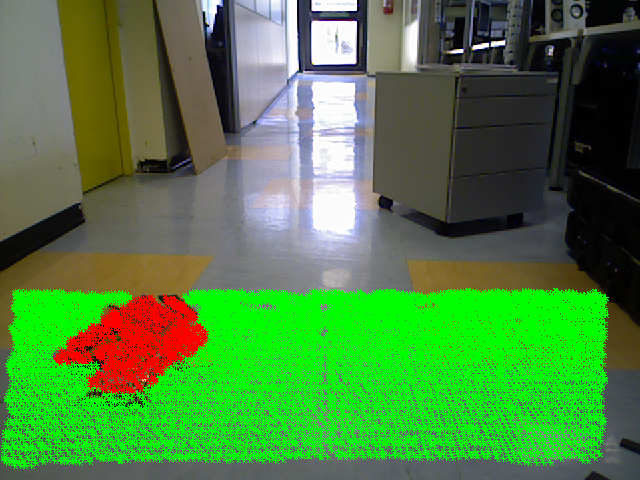}
		\caption{}
		\label{}
	\end{subfigure}\\
\begin{subfigure}{0.23\textwidth}
		\centering
		\includegraphics[width=0.95\textwidth]{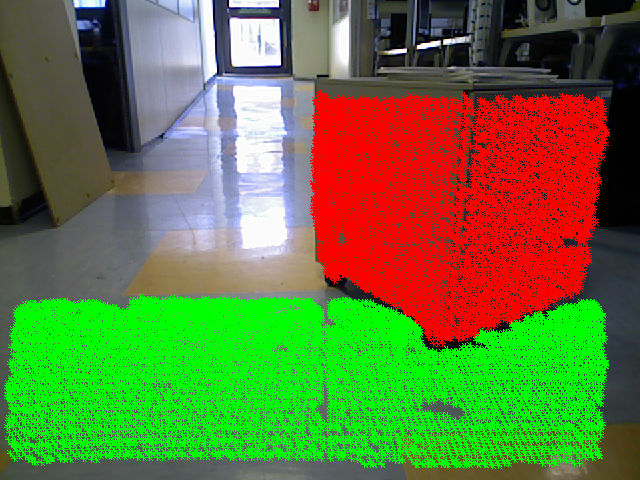}
		\caption{}
		\label{}
	\end{subfigure}
	\begin{subfigure}{0.23\textwidth}
		\centering
		\includegraphics[width=0.95\textwidth]{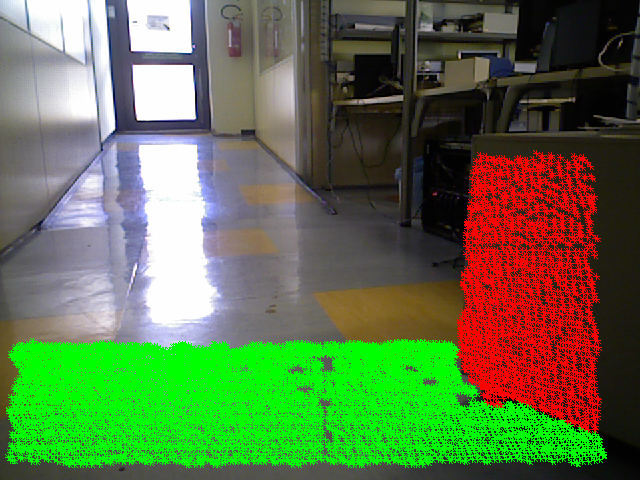}
		\caption{}
		\label{}
	\end{subfigure}
       \caption{Images displaying the path crossed, in sequence from (a) to (f). We projected on them the clusters coloured in function of the traversability; we used a color map where green means \textit{mobility score=1} and red means \textit{mobility score=0}. }
\label{fig:obstaclephotos}
\end{figure}

\begin{figure}[h]
	\begin{subfigure}{0.23\textwidth}
		\centering
		\includegraphics[width=0.8\textwidth]{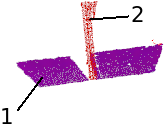}
		\caption{}
		\label{}
	\end{subfigure}
	\begin{subfigure}{0.23\textwidth}
		\centering
		\includegraphics[width=0.8\textwidth]{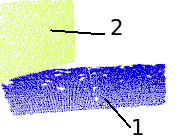}
		\caption{}
		\label{}
	\end{subfigure}\\
	\begin{subfigure}{0.23\textwidth}
		\centering
		\includegraphics[width=0.8\textwidth]{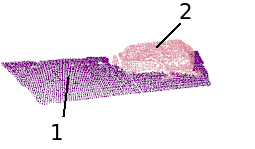}
		\caption{}
		\label{}
	\end{subfigure}
	\begin{subfigure}{0.23\textwidth}
		\centering
		\includegraphics[width=0.8\textwidth]{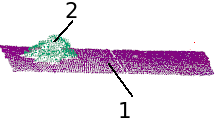}
		\caption{}
		\label{}
	\end{subfigure}\\
	\begin{subfigure}{0.23\textwidth}
		\centering
		\includegraphics[width=0.8\textwidth]{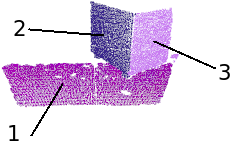}
		\caption{}
		\label{}
	\end{subfigure}
	\begin{subfigure}{0.23\textwidth}
		\centering
		\includegraphics[width=0.8\textwidth]{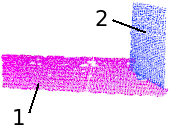}
		\caption{}
		\label{}
	\end{subfigure}
        \caption{Figures presenting point clouds acquired and segmented with our normals/color based method.} 
\label{fig:obclus}
\end{figure} 

Here, we present the results of the usage of our system in conditions more challenging than the one we talked about previously. Fig. \ref{fig:obstaclephotos} shows the projections of the computed mobility maps into the images, while Fig \ref{fig:obclus} shows the segmented point-clouds data.




\subsection{Processing Requirements}
The dataset was processed on an Intel Core i5 CPU@2.30GHz$\times$4. In table \ref{tab:tableperf} we report the time required for each step of the algorithm measured on a sample cloud. Notice that the variations in the number of points processed at each step is due to modifications to the point-cloud done by the previous step, including sampling, outliers removal, and groud separation. This analysis demonstrates that the algorithm can be used to analyse a scene at the rate of two frames per second, which is suitable for foot step planning.
\begin{table}[h!]
\centering  
\begin{adjustbox}{width=0.48\textwidth}
    \begin{tabular}{lrr}
      \toprule
     Step & Time(ms) & Points processed \\
      \midrule
      Denoise & 68.133  & 307200 \\
      Data reduction  &  39.873 & 307200\\
      Normal estimation &  144.766  & 12004\\
	  RGB-N coding & 0.203  & 12004 \\
      Ground removal  & 31.482 & 12004\\
      Color segmentation &  24.961  & 887\\      
	  Surface properties estimation & 175.303  & 887\\
      Mobility mapping  & 1.328 & 887\\
           
      \bottomrule
      Total: & 486.049 &
    \end{tabular}
\end{adjustbox}
\caption{Time performance of our algorithm for analysing one sample point cloud.}
    \label{tab:tableperf}
\end{table}

\section{Conclusion and Future work}
This paper has presented an approach for scene segmentation and mobility estimation based on the analysis of 3D data acquired by a RGBD sensor.
The method was used to successfully and efficiently classify real world indoor workspaces into traversable and untraversable regions. We have achieved a mobility map computation pipeline that runs at 2Hz. Our future work will investigate removing many existing assumptions, including our ground removal algorithm which assumes that the largest segment in any depth map corresponds to the ground. We will also be looking at integrating vision with the depth data to build more robust features for segmentation.





\bibliographystyle{IEEEtran}
\bibliography{IEEEabrv,biblio}

\end{document}